\documentclass{bmvc2k}
\usepackage{float}
\usepackage{caption}
\usepackage{multirow}
\makeatletter
\newcommand*\bigcdot{\mathpalette\bigcdot@{.5}}
\newcommand*\bigcdot@[2]{\mathbin{\vcenter{\hbox{\scalebox{#2}{$\m@th#1\bullet$}}}}}
\makeatother

\title{OriNet: A Fully Convolutional Network for 3D Human Pose Estimation}

\addauthor{Chenxu Luo}{chenxuluo@jhu.edu}{1}
\addauthor{Xiao Chu}{chuxiao@baidu.com}{2}
\addauthor{Alan Yuille}{ayuille1@jhu.edu}{1}

\addinstitution{
 Department of Computer Science\\
 The Johns Hopkins University\\
 Baltimore, MD 21218, USA
}
\addinstitution{
Baidu Research (USA)\\
Sunnyvale, CA 94089, USA
}

\runninghead{Luo \etal}{OriNet: A Fully Convolutional Network for 3D Human Pose}

\def\eg{\emph{e.g}\bmvaOneDot}

\def\etal{\emph{et al}\bmvaOneDot}

\begin{document}

\maketitle

\begin{abstract}
In this paper, we propose a fully convolutional network for 3D human pose estimation from monocular images. We use limb orientations as a new way to represent 3D poses and bind the orientation together with the bounding box of each limb region to better associate images and predictions. The 3D orientations are modeled jointly with 2D keypoint detections. Without additional constraints, this simple method can achieve good results on several large-scale benchmarks. Further experiments show that our method can generalize well to novel scenes and is robust to inaccurate bounding boxes.
\end{abstract}

\section{Introduction}

Estimating 3D human pose from a single RGB image is a fundamental yet challenging problem. It is potentially useful in many real world applications, such as human-robot interaction, augmented reality and character control. 
Besides the inherent challenges in 2D pose estimation, 3D pose estimation from monocular images is considered more difficult due to the loss of depth and scale ambiguity.

With the advent of deep neural networks, we saw significant progress in monocular 3D human pose estimation. Currently, the end-to-end training method~\cite{zhou2017weakly,sun2017compositional} achieve superior results on standard benchmarks~\cite{h36m_pami}. However, there are several limitations that hinder them to real world settings. First, due to the scale and depth ambiguity from a single image, some works~\cite{zhou2017weakly,martinez2017simple} even require a fixed image scale, making it less flexible to generalize to other datasets. 
 
Second, currently most methods require a tightly cropped box around the subject. One reason is that most methods use a fully-connected layer to regress the joint coordinates directly, which makes the network sensitive to backgrounds. Another practical issue is the lack of diverse training data as for 2D human pose estimation~\cite{andriluka14cvpr,lin2014microsoft}.

\begin{figure}[!t]
	\begin{center}
		\includegraphics[width=0.9\linewidth]{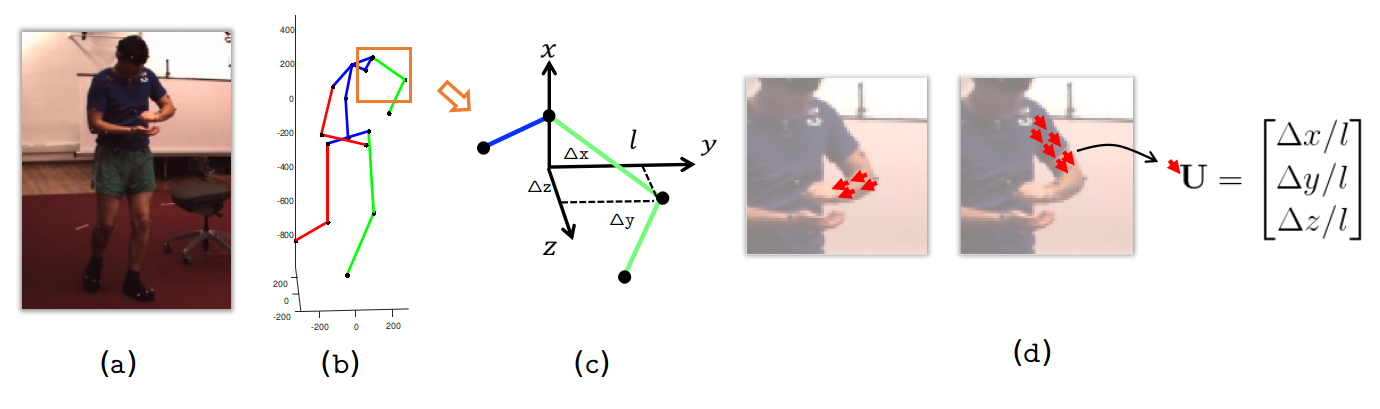}
	\end{center}
	\caption{Modeling limb orientations. 
		(a) is an input example. 
		(b) is the 3D human pose represented with skeleton. The key-points are connected following the tree structure,
		(c) zooming into to see the orientation of the upper arm. 
		$\Delta x,\Delta y$ and $\Delta z$ are the relative coordinate of the two keypoints, $l$ is the length of the limb.
		(d) showing two examples where orientation vectors are binding with the segmentation of the limb.
	}
	\label{fig:firstpage}
\end{figure}
Recently, VNect~\cite{VNect_SIGGRAPH2017} shows promising results of fully convolutional network for 3D human pose estimation. In that work, they attach the 3D joint coordinates to the corresponding locations in the image. However, the performance is not as good as its fully-connected counterparts. One reason may lies on that the network need to localized another joint (the root joint) when predicting the coordinates of each joint. Given that the actual receptive fields of an FCN are much less than the fully connected networks. This makes it harder to regress 3D coordinates of joints far away from the root joint. 

In this paper, we propose a novel fully convolutional network to address the above issues. Instead of directly dealing with joint coordinates, we model limb orientations as a new representation for 3D pose reconstruction. One advantage is that orientation is scale invariant and independent of dataset, which helps resolve the scale ambiguity issue and generalize to diverse data. Since the orientations of limbs are what truly differentiate one pose from another, it is natural and explainable to model the orientations of limbs. Also, it allows the network to focus on pose itself. This representation is flexible and is especially useful for applications such as character control and motion retargeting. Another motivation for using only orientation is that the limb length ratios of different subjects are very similar and are often used as an regularization~\cite{wang2014robust,zhou2017weakly}. So decoupling them to the post-processing stage can let the network focus on each limb without the need to consider other limbs that may lie far away. In our experiments, we show that this is one of the key components to achieve good performance and generalize well to other datasets.

Instead of using a vector representation that lacks spatial association between images and predictions, we propose to combine the orientation with the approximate bounding box of limbs, as shown in Fig.\ref{fig:firstpage} (d). Unlike semantic part segmentation~\cite{xia2016zoom,gong2017look}, we do not care about the detailed boundary of each body part, so a bounding box is a good enough approximate representation for each limb. 
On the orientation map, regions corresponding to limbs are filled with an orientation vectors,
while other locations are set to zero to indicate background. The bounding box of each limb can be easily obtained from 2D joints annotations. This representation can preserve the spatial layout of each limb and explicitly tells the network where to focus when predicting each limb orientation. 

Generally, the proposed method can be plugged into any networks for 2D pose estimation. In this paper, we adopt the Stacked Hourglass network~\cite{newell2016stacked} because of its superior performance. We simply consider the orientations between adjacent joints without in-cooperating additional prior knowledge, we show that our method can still achieve the comparable results on Human 3.6m dataset~\cite{h36m_pami} and start-of-the-art results on MPI-INF-3DHP dataset\cite{mono-3dhp2017}. We expect further constraints such as bone structures~\cite{sun2017compositional,mono-3dhp2017,fang2018learning} or joint-angle limits ~\cite{akhter2015pose} can improve the performance, but they are beyond the scope of this paper.

Our contributions can be summarized as follow:  (1) We propose a fully convolutional network for 3D human pose estimation, which is less sensitive to backgrounds and inaccurate bounding boxes. (2) We propose to use 3D orientations of limbs as a new way to represent 3D poses, which is arguably natural and interpretable. (3) Our proposed method, by only using limb orientations, achieves state-of-the-art results and can generalize well to novel scenes.

\section{Related Work}
Given the difficulties of estimating 3D pose from a single image, many works  decouple the problem into two steps: first estimate 2D joints and then lift then into 3D. 
A typical approach uses sparse-based representation~\cite{ramakrishna2012reconstructing,zhou20153d}. Recent works also use deep network to regress 3D poses from 2D positions directly~\cite{martinez2017simple}.
Thanks to the deep networks, we see huge improvements on 2D pose estimation in recent years~\cite{toshev2014deeppose,wei2016convolutional,chu2016structured,newell2016stacked,chu2016crf,chu2017multi,cao2017realtime}. The progress also benefits such two-stage methods~\cite{zhou2017monocap}. However, these two-step methods rely highly on the results of 2D estimation. Also, discarding image cues makes the problem ill-posed.

Recently, some works try to combine the two steps together. In ~\cite{zhou2016sparseness}, they treat the 2D estimation as latent variables and use the EM algorithm to update the 2D joints at the same time. Tome \etal~\cite{tome2017lifting} refine both the 2D and the 3D estimation iteratively at each stage. However, 3D poses are still estimated merely from the intermediate 2D results.

Recent works ~\cite{zhou2017weakly,sun2017compositional,tekin2016fusing} combine the 2D heatmaps and image cues. This method also needs a lot of training data. Since most of the current datasets only contain indoor scene with limited number of subject and background, some methods~\cite{zhou2017weakly,sun2017compositional} mix 2D and 3D data during the training time. This makes it less flexible and often need manually correct the discrepancy~\cite{zhou2017weakly} between dataset annotations. ~\cite{pavlakos2018ordinal,relativeposeBMVC18} use ordinal depth as additional supervision.

Most of the above methods use 3D coordinates to represent 3D poses. The coordinates can be relative positions to a root joint~\cite{tome2017lifting,martinez2017simple,VNect_SIGGRAPH2017},to the adjacent joint~\cite{li20143d} or their combinations~\cite{mono-3dhp2017}. Another way is to use 2D pixel coordinates and depth to represent each joint~\cite{pavlakos2017coarse,sun2017compositional} and assuming known intrinsic parameter to recover the final 3D pose . 

A few works model limbs for pose estimation. In~\cite{sun2017compositional} they define bone errors along the skeleton chain. Different from theirs, our method only consider each limb independently without long-range dependencies, making it more flexible to deal with rare poses.
Zhou \etal~\cite{zhou2016deep} also tries to deal with joint angles. They define a set of joint angles as intermediate representation and use a kinematic layer to reconstruct 3D poses. Here we only use the orientations as output. We show that this simple method can achieve better results.

To best of our knowledge, only two kinds of works use fully convolutional networks for 3D pose estimation. Pavlakos \etal~\cite{pavlakos2017coarse} introduces 3D heatmap to predict per-voxel likelihood by discretizing depth values. But they need the groundtruth depth of a root joint to finally recover the 3D pose, making it less practical. VNect~\cite{VNect_SIGGRAPH2017} attach the 3D coordinates to a neighbor of the joint on the heatmap. 
Both of them only deal with each joint separately, without considering the limb orientation which is the underlying factor that causes different poses. Given the flexibility of FCNs, in this paper we propose a more powerful methods for 3D pose estimation.

The orientation has only been explored in the context of 2D pose~\cite{cao2017realtime} previously. However, it only serves as an auxiliary task to differentiate different instances, so there is no quantitative evaluation for that. Also it is nontrivial whether it can be extended to estimating 3D orientations accurately, for 3D poses need reasoning beyond the image space. The motivation and application are totally different from ours. 

\section{Method}

In this section, we introduce our fully convolutional network for 3D human pose estimation in detail. Our method model the 2D key-point positions together with the orientation of limbs. The orientation is formulated with an approximate bounding box of the corresponding limb region. 
An overview of our framework is shown in Fig. \ref{fig:overview}. 
\subsection{Orientation Representation}
The orientation of each limb is the most discriminative property of 3D poses. It can be represented by a unit vector $\mathbf{U} = (\Delta x,\Delta y,\Delta z)/l$, 
where $\Delta x,\Delta y$ and $\Delta z$ are the relative positions of the two joints.
$l  = \sqrt{\Delta x^2+\Delta y^2+\Delta z^2}$ is the length of the limb.
The normalized orientations remove the influence of different human scales and image resolutions.

In order to preserve this spatial layout and explicitly tell the network where to focus, we propose to model the orientation together with the limbs region, as shown in Fig. \ref{fig:overview}(e).
For each limb $k$, we generate a bounding box using the locations of two endpoint joints correspond to that limb on the label map. 
Specifically, the bounding box region $\mathcal{L}_k$ contains pixels within a predefined width $w_k$ to the line segment ${\mathbf{p}_{k_1}\mathbf{p}_{k_2}}$ of two joints $k_1$ and $k_2$. This forms a good approximation for each limb region, see Fig. \ref{fig:overview}(e) and supplementary material for detail.
Pixels inside the bounding box indicate the region of this limb, and are labeled with the orientation vectors.
Other places are set to zero,indicating background. The orientation map is defined as:
\begin{equation}
\widehat{\mathbf{O}}_k(i,j) = \left\{
\begin{aligned}
& \mathbf{U}_k & (i,j) &\in \mathcal{L}_k, \\
& \mathbf{0} & (i,j) & \notin\mathcal{L}_k,
\end{aligned}
\right.
\label{eq:seglabel}
\end{equation}
where $(i,j)$ are the pixel locations on the output prediction map,
$\mathcal{L}_k$ are the regions corresponding to the limb $k$. Examples of feature map are shown in Fig. \ref{fig:overview}.

The loss function for limb orientations is as follows,
\begin{equation}
\mathrm{L}_{\text{o}} =  \sum_{k} \|\mathbf{O}_k - \widehat{\mathbf{O}}_k \|_2^2 
\end{equation}
where $\mathbf{O}_k$ is the predicted orientation map for the $k$-th limb.

\subsection{Model the 2D keypoint location}
In order to encourage the connection between the 2D appearance and 3D orientations, we propose to predict the 2D keypoint at the same time. The 2D joints can help localize the limb region more precisely during both training and inference. 

Following the common practice, we represent 2D keypoints with heatmaps, as shown in Fig. \ref{fig:overview}. In each heatmap, a Gaussian centered at ground-truth location indicate the existence of that keypoint. 
We train 2D keypoints with sigmoid cross entropy loss:
\begin{equation}
\begin{aligned}
\mathrm{L}_\text{p} = - \frac{1}{N}\sum_{n} \sum_{i=1}^{W}\sum_{j=1}^{H}
\mathbf{p}_{i,j}^n \log(\mathbf{\widehat{p}}_{i,j}^n) +  (1- \mathbf{p}_{i,j}^n )  \log(1-\mathbf{\widehat{p}}_{i,j}^n)
\end{aligned} 
\end{equation}
where $\mathbf{p}^n$ and $\mathbf{\widehat{p}}^n$ are predicted and groundtruth heatmaps and $N$ is the number of joints.

The final loss function is 
\begin{equation}
L = L_o+\lambda L_{\text{p}}
\end{equation}
where $\lambda$ is a balancing parameter. We use $\lambda=0.2$ in the experiments. 

\subsection{Network Structure}
\begin{figure*}[t]
	\begin{center}
		\includegraphics[width=1\linewidth]{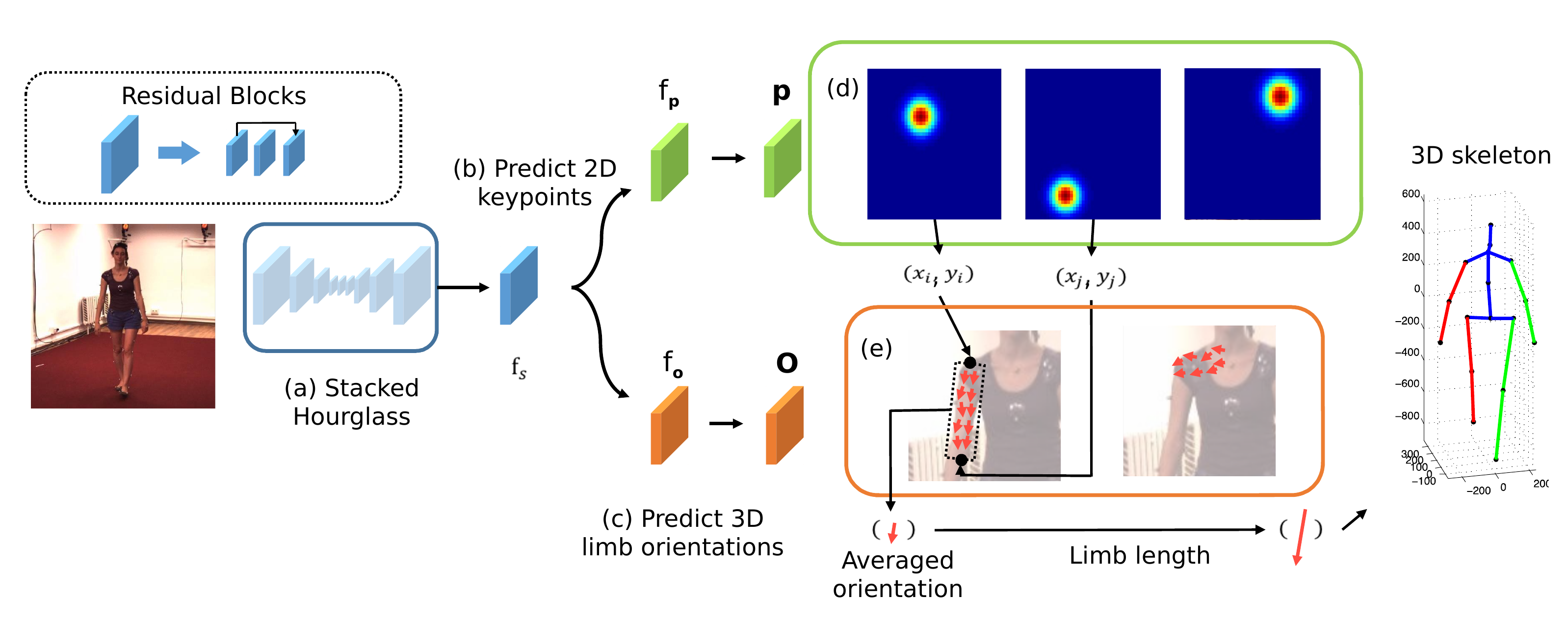}
	\end{center}
	\caption{Overview of our framework. 
		We use both 2D heatmaps (d) and limb orientations (e) as the supervisions. The 3D orientation maps are derived from keypoints without extra annotations.
		In the test stage, the predicted keypoint locations are used to crop the limb regions on the orientation heatmap to get the orientation prediction.}
	\label{fig:overview}
\end{figure*}
\begin{figure*}[t!]
	\begin{center}
		\includegraphics[width=0.8\linewidth]{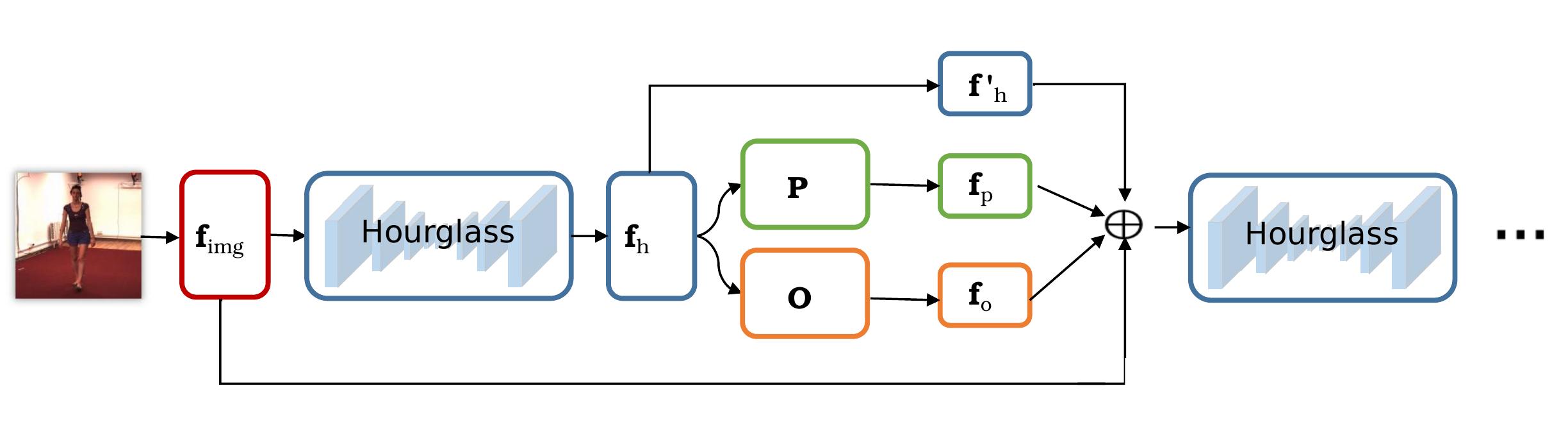}
	\end{center}
	\caption{Network structure. 
		$\mathbf{f}_{\text{h}}$ is the output of the hourglass module,
		$\mathbf{P}$ is the 2D heatmaps
		and $\mathbf{O}$ is orientation maps. $\mathbf{f}_{\text{img}}$ is the image feature and 
		$\mathbf{f}_{\text{p}}$, $\mathbf{f}_{\text{o}}$ and $\mathbf{f}_{\text{h}}^{\prime}$ are the intermediate features,
	}
	
	\label{fig: network structure}
\end{figure*}

We implemented the model based on the Stacked Hourglass network~\cite{newell2016stacked}.
The network learns low level image representations $\mathbf{f}_{\text{img}}$ for the input image.
In each stack, the hourglass module will learn its feature $\mathbf{f}_{\text{h}}$.
After that, the network is split into two branches, one for 2D joint locations, the other for the 3D limb orientations (see Fig. \ref{fig:overview}).
Supervision is applied after each stack.

The predictions, hourglass features and the image features are combined as follows:
\begin{align} 
\mathbf{f}_{\text{img}}^{\prime} & = \mathbf{f}_{\text{img}} + \text{Conv}(\mathbf{p})+ \text{Conv}(\mathbf{O}) +\text{Conv}(\mathbf{f}_{\text{h}})
\end{align}
Then the updated image feature $\mathbf{f}_{\text{img}}^{\prime}$ is used as the input of next hourglass module.
For simplicity, we draw the Hourglass network structure for 2 stacks in Fig. \ref{fig: network structure}.

\subsection{Inference}
The inference process includes 
(1) estimating 2D keypoints,
(2) using the estimated 2D keypoints to read off limb regions, 
(3) taking the averaged orientations in that region,
(4) recovering the 3D pose using the estimated orientations, limb length ratio and scale information.

The 2D keypoint locations can be obtained by finding the maximal location on the 2D key-point heatmap, for the $n$-th joint, 
\begin{equation}
(x_n,y_n)  = \arg\max \mathbf{p}^{n}
\end{equation}

After getting predictions of all 2D keypoints, we can define the region for each limb.
Taking a pair of adjacent joints $\mathbf{p}_1$ and $\mathbf{p}_2$ that corresponds to limb $l$,
we use the coordinates to crop out the region between them on the limb 3D orientation 
$\mathbf{O}_l$ . The cropped regions contain the predictions of the orientation of the $l\text{-th}$ limb.

We average the value of the normalized predictions and then normalize it as the estimated orientation.
Then the limb length ratio as used in many other works~\cite{zhou2017weakly,wang2014robust} and scale information~\cite{zhou2016sparseness,zhou2017weakly,martinez2017simple,sun2017compositional} are used to recover each limb vector. 
By choosing a root joint, we reconstruct each joint position along the tree structure iteratively. Indeed this simple representation could cause error accumulation along the path, however, we show that this can still achieve good results without adding additional constraints.

\section{Experiments}
\subsection{Datasets and Implementation Detail}
We evaluate our method on two datasets: Human3.6m ~\cite{h36m_pami} and MPI-INF-3DHP~\cite{mono-3dhp2017}.\\
\textbf{Human3.6M.} This is currently the largest 3D pose dataset, containing 11 subjects performing 15 actions. We adopt a commonly used protocol: five subjects(1,5,6,7,8) for training and two subjects (9,11) for testing. The videos are down sampled to 10fps. Mean per joint position error(MPJPE) is computed with the root joint aligned. We also report the results after Procrustes alignment, which only focus on the structure of the 3D poses. Following \cite{zhou2016sparseness}, we test our algorithm on all 17 joints defined in \cite{h36m_pami}.\\
\textbf{MPI-INF-3DHP.} This is a newly released dataset. It is more challenging, for it contains more diverse motions and aims for testing the generalization of various methods. Here, we only use the provided training data and do \textbf{not} use any background augmentation like~\cite{mono-3dhp2017,VNect_SIGGRAPH2017}. We report the PCK with a threshold of 150mm and AUC on all 2929 testing images.\\
\textbf{Implementation Detail.}
We adopt a 5-stack hourglass network~\cite{newell2016stacked}  and implement it in Torch7~\cite{collobert2011torch7}. The initial learning rate is 2.5e-4 and decreases by a factor of 10 after every two epochs. We train the model for about 8 epochs, using RMSprop optimizer. We also add scale and color jittering as data augmentation. 
During the testing time, for fair comparison, we assume known bounding box of the person as in all the other works. We only use a single crop, without using flipping or multiple scale fusion. For the scale issue, we use training subjects to compute the limb-length ratio and scale. We also test our model directly with randomly shifted bounding box randomly as in \cite{VNect_SIGGRAPH2017}. Our approach can run at 20fps on a Titan XP.

\subsection{Results on Human3.6M Dataset}
\begin{figure*}
	\begin{center}
		 \begin{tabular}{@{}cc@{}}
    \includegraphics[width=0.45\columnwidth]{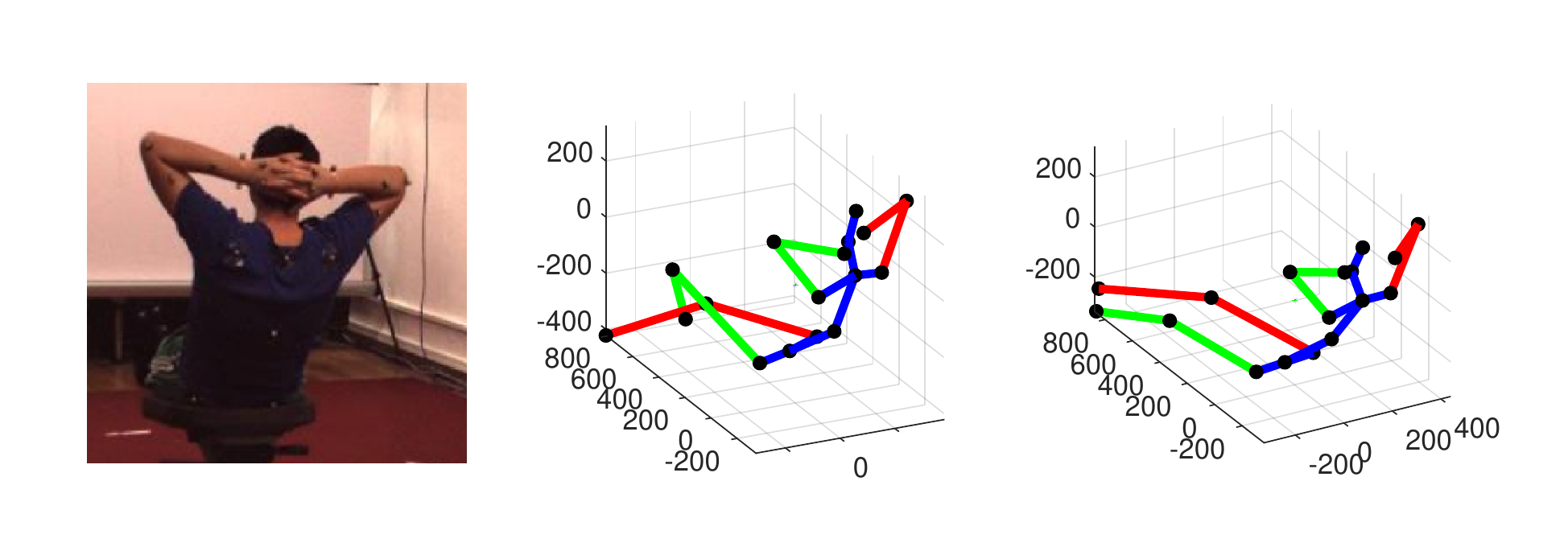} & \includegraphics[width=0.45\columnwidth]{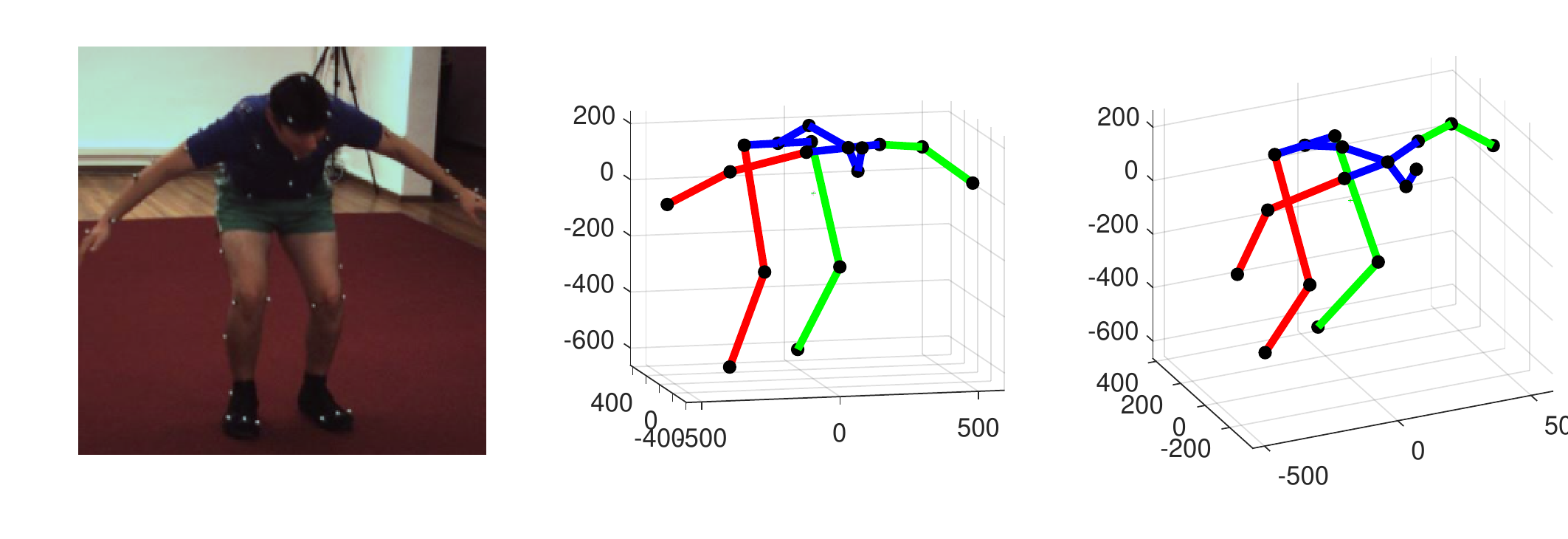}
  \end{tabular}
			\begin{tabular}{@{}cc@{}}
    \includegraphics[width=0.45\columnwidth]{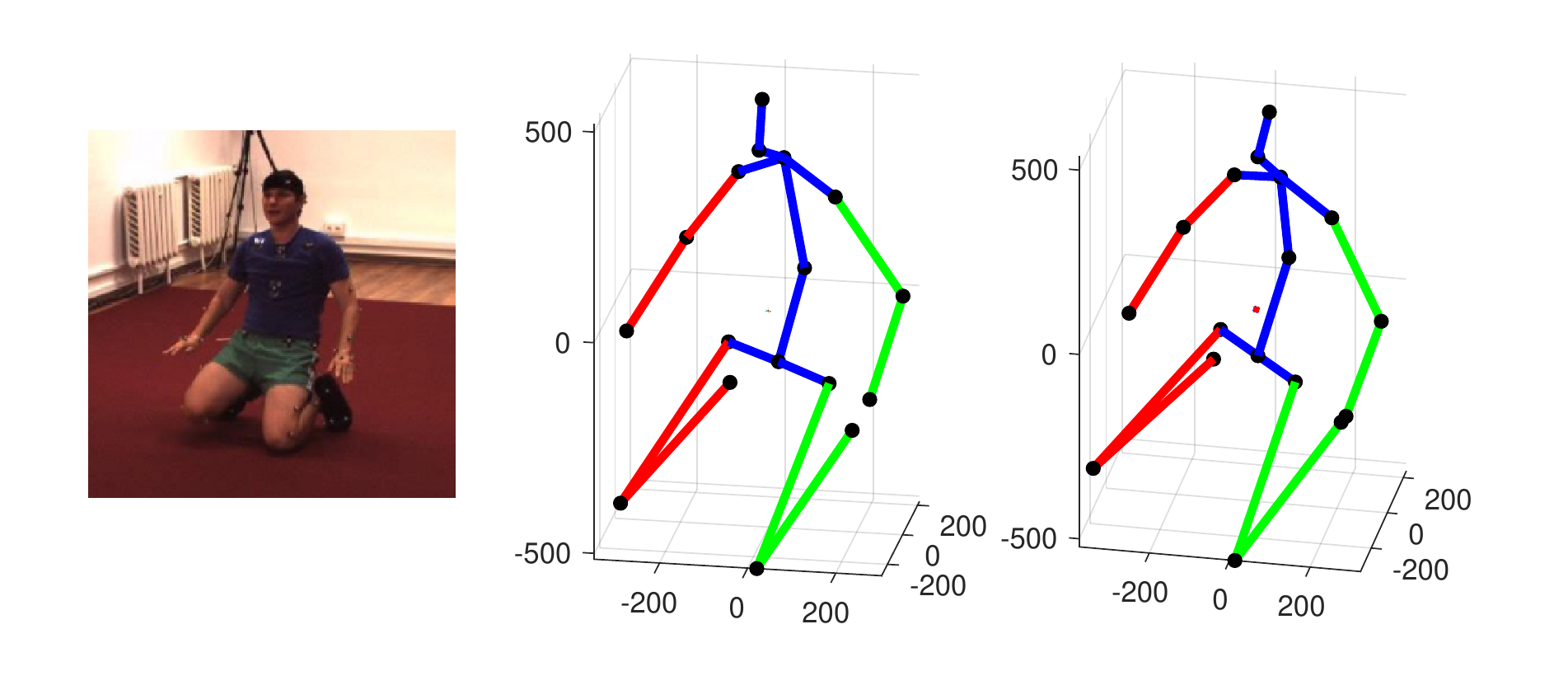} & \includegraphics[width=0.45\columnwidth]{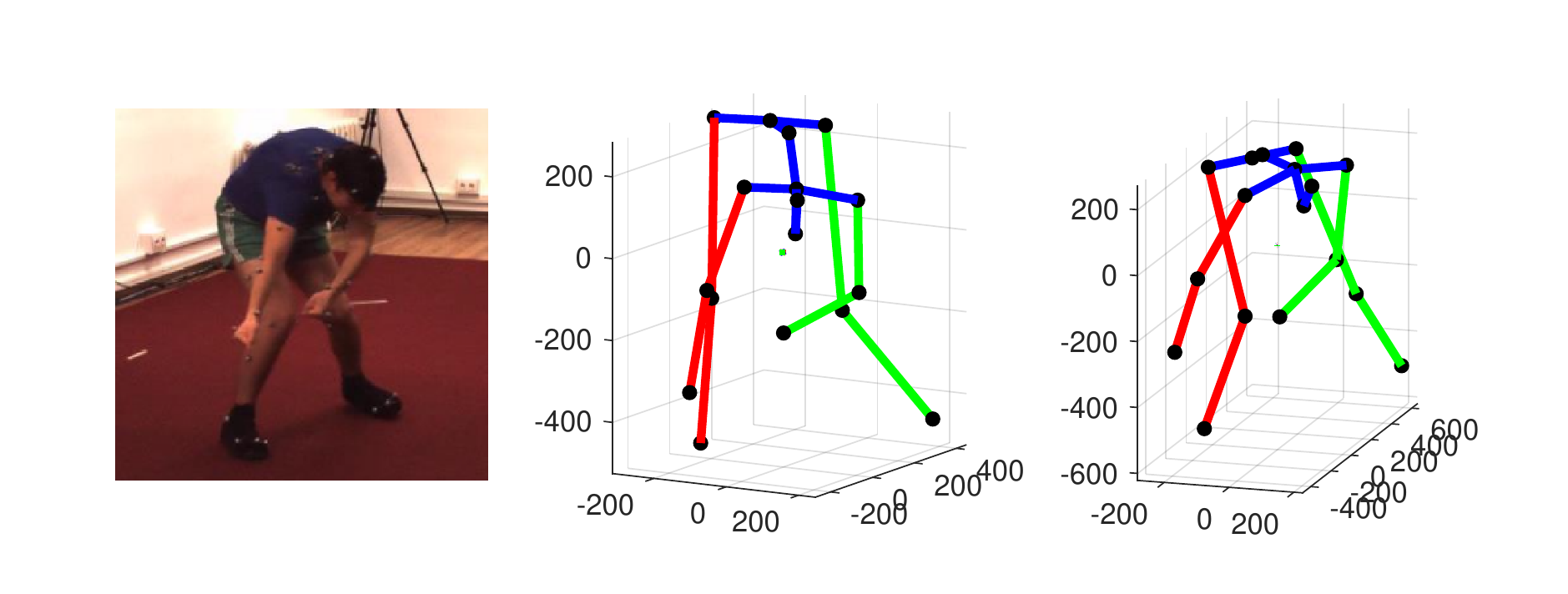}
  \end{tabular}
	\end{center}
	\caption{Qualitative results on Human3.6M dataset. In each example, the first one is the input image, the second one is our estimated pose and the third on is the groundtruth.}
	\label{fig:h36m}
\end{figure*}

Results are shown in Table \ref{table:protocol1}. 
For model trained only on Human3.6m dataset, we achieve the state-of-the-art result. Especially for difficult poses like sitting down, which have fewer examples in the training set, our method outperforms other methods by a large margin. This shows that our method is more data efficient. As for models pretrained or mixed trained with 2D datasets (\eg MPII), our method can still achieve comparable results. Compared with ~\cite{sun2017compositional,martinez2017simple}, our method has lower error after alignments while slightly higher errors before alignment. Note that ~\cite{sun2017compositional} also use 2D data during training and camera parameters during testing. The result shows that our predicted poses have the most similar structure compared with the groundtruth. We also report results using the groundtruth limb length, similar to the skeleton-fitting step used in VNect~\cite{VNect_SIGGRAPH2017}.
\begin{table*}[h]
\small
	\begin{center}
		\tabcolsep=0.02cm
		\begin{tabular}{|l|c|c|c|c|c|c|c|c|c|c|c|c|c|c|c|c|}
			\hline
			& Direct & Diss. & Eat & Greet & Phone &Photo & Pose & Purch.  & Sit & SitD. & Smoke & Wait & WalkD & Walk & WalkT & \textbf{Ave} \\
			\hline\hline
			\multicolumn{17}{|l|}{Train on Human3.6M from scratch(or pretrained on ImageNet~\cite{Russakovsky2015})} \\ \hline
			\cite{sun2017compositional} & 90.2 & 95.5 & 82.3 & 85.0 & 87.1 & 94.5 & 87.9 & 93.4 & 100.3 & 135.4 & 91.4 &  87.3 & 90.4 & 78.0 & 86.5 & 92.4\\
			\cite{tome2017lifting} & \textbf{64.9}  & \textbf{73.5} &  76.8  & 86.4  & 86.3 &  110.7 &  \textbf{68.9}  & \textbf{74.8}& 110.2 &  173.9 &  84.9 &  85.8  & 86.3  & 71.4 &  73.1  & 88.4 \\
			Ours & 68.4 & 77.3 & \textbf{70.2} & \textbf{71.4} &\textbf{75.1} & \textbf{86.5} &69.0 & 76.7 & \textbf{88.2} & \textbf{103.4} & \textbf{73.8} & \textbf{72.1} & \textbf{83.9} & \textbf{58.1} & \textbf{65.4} & \textbf{76.0}\\ \hline
			\multicolumn{17}{|l|}{Pretrained on 2D pose datasets(e.g. MPII)} \\ \hline
			\cite{mono-3dhp2017} & 59.7 & 69.7 & 68.8 & 68.8 & 76.4 & 85.2 & 59.0 & 75.0& 96.2 & 122.9 & 70.8 & 68.5 & 82.0 & 54.4 & 59.8 & 74.1 \\
			\cite{tekin2016fusing} & 54.2  & 61.4 & 60.2 & 61.2 &  79.4 & 78.3 &  63.1 &  81.6 &  \textbf{70.1} & 107.3 & 69.3 & 70.3 & 74.3 & 51.8  & 63.2 & 69.7\\
 \cite{martinez2017simple} & \textbf{51.8} &\textbf{56.2} & 58.1 & \textbf{59.0} & 69.5 & 78.4 & \textbf{55.2} & \textbf{58.1} & 74.0 & \textbf{94.6} & \textbf{62.3} & 59.1 & \textbf{65.1} &  49.5 & 52.4 & \textbf{62.9}\\
			Ours & 53.5 & 60.9 & \textbf{56.3} & 59.1 & \textbf{64.3} & \textbf{74.4} & 55.4 & 63.4 & 74.8 & 98.0 & 61.1 & \textbf{58.2} & 70.6 & \textbf{49.1} & 55.7 & 63.7\\ \hline
			Ours$^*$ & 49.2 & 57.5 & 53.9 & 55.4 & 62.2 & 73.9 & 52.1 & 60.9 & 73.8 & 96.5 & 60.4 & 55.6 & 69.5 & 46.6 & 52.4 & 61.3\\ \hline \hline
            \multicolumn{17}{|l|}{Results after Procrustes alignment} \\ \hline
            \cite{moreno20163d} & 69.5& 80.2 &78.2 &87.0 &100.7&102.7& 76.0& 69.6 &104.7& 113.9& 89.7  &98.5& 79.2& 82.4 &77.2 & 87.3\\
            \cite{sun2017compositional} & 42.1 & 44.3 & 45.0 & \textbf{45.4} & 51.5 & \textbf{53.0} & 43.2 & 41.3 & 59.3 & 73.3 & 51.0 & 44.0 & 48.0 & 38.3 & 44.8 & 48.3 \\
            \cite{martinez2017simple} & \textbf{39.5} & \textbf{43.2}&  46.4&  47.0&  51.0 & 56.0&  \textbf{41.4} &  \textbf{40.6} &  56.5  &  \textbf{69.4} & 49.2  &45.0  &49.5  &38.0  &43.1 & 47.7 \\
			Ours & 40.8 & 44.6 & \textbf{42.1} & 45.1 & \textbf{48.3} & 54.6 & 41.2 & 42.9 & \textbf{55.5} &69.9 & \textbf{46.7} & \textbf{42.5} & \textbf{48.0} & \textbf{36.0} & \textbf{41.4} & \textbf{46.6} \\ \hline
			
		\end{tabular}
	\end{center}
	\caption{Mean per joint position errors (MPJPE) in mm on Human3.6M. $*$ denotes using the groundtruth length for each limb.} 
	\label{table:protocol1}
\end{table*}
In Fig. \ref{fig:h36m}, we show some qualitative results from Human3.6M dataset. For each example, we illustrate the image, our result and the groundtruth respectively. 

\subsection{Results on MPI-INF-3DHP}
\subsubsection{Cross Datasets Test}
Performance on Human3.6M seems to be saturating. However, the generalization problem largely remain unexplored. MPI-INF-3DHP provides a good testbed for such purpose. We apply models trained on Human3.6m directly to this dataset and results are shown in table \ref{table:mpi-inf-3dhp_h36m}. 

\begin{table}
\small
	\begin{center}
		\tabcolsep=0.11cm
		\begin{tabular}{|c|c|c|c|c|c|c|c|}
			\hline
			\multirow{2}{*}{Method} & \multirow{2}{*}{Training Data} & \multicolumn{4}{c|}{PCK}    & AUC      \\  \cline{3-7} 
			&   & GS   & noGS    & Outdoor   & ALL     & ALL       \\ \hline
			Meta~\cite{mono-3dhp2017}  & H36m & 70.8 & 62.3 & 58.5 & 64.7 & 31.7 \\
			Pavlakos~\cite{pavlakos2018ordinal} & H36m & - & - & - & 17.1 & 6.3 \\
			Zhou~\cite{zhou2017weakly} & H36m &45.6 & 45.1 & 14.4 & 37.7 & 20.9 \\
			Martinez$^*$~\cite{martinez2017simple} & H36m &49.8 & 42.5 & 31.2 & 42.5 & 17.0 \\
            Ours-len$^*$ & H36m & 65.7 & 56.0 & 60.3 & 60.9 & 29.9 \\
            Our & H36m & 69.8 & 58.3 & 65.5 & 64.6 & 32.1  \\
			Ours$^*$  & H36m &\textbf{71.3} & 59.4 & \underline{65.7}  &\underline{65.6}  & \textbf{33.2} \\ \hline \hline
            Zhou~\cite{zhou2017weakly} & H36m+MPII & \underline{71.1}          & \textbf{64.7}          & \textbf{72.7}          & \textbf{69.2}          & \underline{32.5}         \\ 
            Pavlakos~\cite{pavlakos2018ordinal} & H36m+MPII & - & - & - & 44.3 & 19.8 \\ 
            Pavlakos~\cite{pavlakos2018ordinal} & H36m+MPII+Ord & 76.5 & 63.1 & 77.5 & 71.9 & 35.3 \\ \hline
            
		\end{tabular}
	\end{center}
	\caption{Results on MPI-INF-3DHP test set by scene. All models are pretrained on MPII. Our method can achieve comparable results even with mixed training method~\cite{zhou2017weakly} .}
	\label{table:mpi-inf-3dhp_h36m}
\end{table}

Our method can achieve comparable results with~\cite{mono-3dhp2017}, which is specifically designed for transfer learning. Even compared with mixed-training method~\cite{zhou2017weakly}, we still get comparable results. However, as pointed out in~\cite{zhou2017weakly}, they need to manually fix the position of some joints in an ad-hoc manner because of the discrepancy between different annotations.  

We also test the direct regression method~\cite{martinez2017simple} using 2D groundtruth as input and rescale the output to the universal skeleton. We see that it can not generalize well to different camera viewpoints. Our method can deal with different camera viewpoints and background better. 

We also test the model trained on Human3.6m that uses relative limb vector (to the length of the torso) instead of orientations. During the testing time, we rescaled each limb to its groundtruth length, only preserving the orientations. We show that it generalize worse than using orientation as representation.

\subsubsection{Training on MPI-INF-3DHP Dataset}

We finetune the network trained on the Human3.6m previous on this new dataset, without adding background augmentations. Results are shown in Table \ref{table:mpi-inf-3dhp}. Compared with ~\cite{mono-3dhp2017,VNect_SIGGRAPH2017}, which use extensive background and clothes augmentation, our model trained merely on the green-screen background can still outperform other methods. This shows that our method is more robust to background and can transfer well to different scenes.

\begin{table*}
\small
	\centering
    \tabcolsep=0.05cm
	\begin{tabular}{l|l|l|l|l|l|l|l|l|l|l|l|}
		& data                                                                  & Walk & Exe.              & \begin{tabular}[c]{@{}l@{}}Sit\\ \end{tabular} &  Reach & Floor & Sport                 & Misc                  & \multicolumn{3}{c|}{Total}                             \\ \hline
		&                                                                             & PCK                                                   & PCK                    & PCK                                                    & PCK                                                     & PCK                                                    & PCK                    & PCK                    & PCK           & AUC           & MPJPE              \\ \hline
		Mehta~\cite{mono-3dhp2017} & \begin{tabular}[c]{@{}l@{}}(MPII)\\ 3DHP$^a$\\ \qquad +H3.6M \end{tabular}                  &\begin{tabular}[c]{@{}c@{}} \\ 86.6\\ -      \end{tabular}                                            & \begin{tabular}[c]{@{}c@{}} \\75.3\\ -      \end{tabular}                    & \begin{tabular}[c]{@{}c@{}} \\ 74.8 \\ -      \end{tabular}                                                   &  \begin{tabular}[c]{@{}c@{}} \\73.7   \\ -      \end{tabular}                                                  &  \begin{tabular}[c]{@{}c@{}} \\52.2\\ -      \end{tabular}                                                    &  \begin{tabular}[c]{@{}c@{}} \\82.1 \\ -      \end{tabular}                   &  \begin{tabular}[c]{@{}c@{}} \\77.5  \\ -      \end{tabular}                  &  \begin{tabular}[c]{@{}c@{}} \\75.7 \\ 76.5     \end{tabular}          &  \begin{tabular}[c]{@{}c@{}} \\39.3 \\ 40.8      \end{tabular}          &  \begin{tabular}[c]{@{}c@{}} \\117.6   \\ -      \end{tabular}                \\ \hline
  Mehta~\cite{VNect_SIGGRAPH2017} & \begin{tabular}[c]{@{}l@{}}(MPII+LSP)\\ H3.6M+3DHP$^a$ \end{tabular}       
  & \begin{tabular}[l]{@{}l@{}}  87.7  \end{tabular}
  & \begin{tabular}[l]{@{}l@{}} 77.4  \end{tabular}
  & \begin{tabular}[l]{@{}l@{}} 74.7  \end{tabular}
  & \begin{tabular}[l]{@{}l@{}} 72.9  \end{tabular}
  & \begin{tabular}[l]{@{}l@{}} 51.3  \end{tabular}
  & \begin{tabular}[l]{@{}l@{}} 83.3  \end{tabular}
  & \begin{tabular}[l]{@{}l@{}} 80.1  \end{tabular}
  & \begin{tabular}[l]{@{}l@{}} 76.6  \end{tabular}
  & \begin{tabular}[l]{@{}l@{}} 40.4 \end{tabular}
  & \begin{tabular}[l]{@{}l@{}} 124.7 \end{tabular}
        \\ \hline
Dabral~\cite{Dabral_2018_ECCV} & \begin{tabular}[c]{@{}l@{}}MPII+\\3DHP$^a$ \end{tabular}       
& \begin{tabular}[l]{@{}l@{}}  89.1$^*$  \end{tabular}
& \begin{tabular}[l]{@{}l@{}} 75.1$^*$  \end{tabular}
& \begin{tabular}[l]{@{}l@{}} 73.6$^*$  \end{tabular}
& \begin{tabular}[l]{@{}l@{}} 77.9$^*$  \end{tabular}
& \begin{tabular}[l]{@{}l@{}} 49.2$^*$  \end{tabular}
& \begin{tabular}[l]{@{}l@{}} 79.3$^*$  \end{tabular}
& \begin{tabular}[l]{@{}l@{}} 80.8$^*$  \end{tabular}
& \begin{tabular}[l]{@{}l@{}} 76.7$^*$  \end{tabular}
& \begin{tabular}[l]{@{}l@{}} 39.1$^*$ \end{tabular}
& \begin{tabular}[l]{@{}l@{}} 103.8$^*$ \end{tabular}
\\ \hline	
		Ours  & \begin{tabular}[c]{@{}l@{}}(MPII)\\ H3.6M+3DHP\end{tabular}           & \begin{tabular}[l]{@{}l@{}} \textbf{90.4} \\ 90.5$^*$   \end{tabular}             & \begin{tabular}[l]{@{}l@{}} \textbf{79.1}  \\ 80.9$^*$  \end{tabular}             &  \begin{tabular}[l]{@{}l@{}} \textbf{88.5} \\ 90.0$^*$   \end{tabular}       
  &  \begin{tabular}[l]{@{}l@{}} \textbf{81.6}\\ 85.6$^*$   \end{tabular}         
  &  \begin{tabular}[l]{@{}l@{}} \textbf{66.3}\\ 70.2$^*$   \end{tabular}         
  &  \begin{tabular}[l]{@{}l@{}} \textbf{91.9}\\ 93.0$^*$   \end{tabular}           
  &  \begin{tabular}[l]{@{}l@{}} \textbf{92.2}\\ 92.9$^*$   \end{tabular}           
  &  \begin{tabular}[l]{@{}l@{}} \textbf{81.8}\\ 83.8$^*$   \end{tabular}     
  &  \begin{tabular}[l]{@{}l@{}} \textbf{45.2}\\ 47.7$^*$   \end{tabular}     
  &  \begin{tabular}[l]{@{}l@{}} \textbf{89.4}\\ 85.0$^*$   \end{tabular}             \\ \hline
 
	\end{tabular}
\vspace{3mm}
\caption{Activitywise performance on MPI-INF-3DHP test set. We report the PCK, AUC and MPJPE. Higher PCK and AUC are better and lower MPJPE is better. (MPII) means pretrained on MPII dataset.``$a$'' means adding background augmentation for training, ``$p$'' means Procrustes alignment, $*$ denote using groundtruth limb length.}
\label{table:mpi-inf-3dhp}
\end{table*}
\subsection{Ablation Study}
In this part, we evaluate different components of our approach on the Human3.6m. For simplicity, we use another protocol used in~\cite{yasin2016dual,rogez2016mocap,chen20163d}, where only one subject (S11) is used for testing. The estimated 3D pose is first aligned with a rigid transformation, which ruling out factors such as scale and rotation. We trained the 1-stack networks from scratch and 5-stack networks pretrained on the MPII dataset. Results are shown in table \ref{table:protocol2}.
\begin{table*}[h]
	\begin{center}
		\tabcolsep=0.07cm
		\begin{tabular}{|l|c||l|c|}
			\hline
   method & MPJPE & method & MPJPE \\ \hline
Chen\cite{chen20163d} &  82.72 &   Chen\cite{chen20163d}+2D GT  & 57.50 \\ \hline
			1-stack &  50.68 & 1-stack-fc & 56.09 \\
			1-stack-len  &  86.32 & 1-stack-len-fc & 81.51 \\ \hline
            5-stack-len & 52.72 & 5-stack-len(rescaled) &41.95 \\
			5stack &  \textbf{38.53} & & \\ \hline
		\end{tabular}
	\end{center}
	\caption{The mean reconstruction errors on Human3.6M (S11). Errors are computed after Procrustes alignment. ``fc'' means sparse representation with a fully connected layer at the end. ``len'' means using the properly normalized limb vector instead of orientations. ``Rescaled'' means rescaled each bone to the groundtruth length during testing.}
	\label{table:protocol2}
\end{table*}
\subsubsection{Orientation}
First, we demonstrate the advantage of estimating orientations compared with the original limb vector. 
We train the same network using the original limb vector properly normalized by the length of torso.  As we can see, using the limb orientation representation can perform significantly better than considering bone length at the same time for both 1-stack and 5-stack networks (see 1-stack-len vs 1-stack and 5-stack-len vs 5-stack respectively). Even if we rescaled the bone vector to the groundtruth length preserving the orientation, the result is still worse than directly regressing the orientation. Furthermore, the 1-stack network using the orientation representation can even outperform 5-stack network with the original limb vector. This shows that decoupling the orientation and length of each limb is beneficial.
\subsubsection{Image to Prediction Association}
In this part, we show the advantages of having the spatial associations between images and predictions.
We train 1-stack networks from scratch, one with dense output as proposed, one with max pooling layer after the last several convolutional layers and the final convolutional layer replaced by a fully connected layer for direct regression. The dense representations achieves better results using orientations as representation. This show the benefits of our proposed image-to-prediction association. However, by considering limb length, the fully convolutional does not perform as well as the fully connected network. We conjecture  that it may due to the relative small receptive field the network actually has. It also shows the necessary for using orientation for fully convolutional networks to achieve better performance.


\subsubsection{Robust to Bounding Box Jitter and Scale}
Similar to ~\cite{VNect_SIGGRAPH2017}, we carry out experiments on MPI-INF-3DHP dataset by jittering the bounding box at random in the range of $\pm40$ px. Since during the testing, we need to resize the bounding boxes to a fixed scale(256 in our experiments and 224 in~\cite{VNect_SIGGRAPH2017}), we also add more noises by jittering the bounding box in the range of $\pm100$ px and rescale it by a factor of $\pm 0.2$. Our method is less sensitive to inaccurate bounding boxes. \\
\begin{center}
\small
\begin{tabular}{{|l|c|c|}}
\hline
& PCK & AUC \\ \hline
Mehta~\cite{VNect_SIGGRAPH2017}(40px) &  70.1($\downarrow\sim6$) & 35.7 \\ \hline
ours(40px) & 80.8$\pm$0.16 ($\downarrow\sim1)$ & 44.3$\pm$0.11\\ 
ours(100px+rescale) & 77.9 ($\downarrow\sim4$) & 41.3 ($\downarrow\sim4$) \\ \hline
\end{tabular}
\end{center}



\begin{figure*}
	\begin{center}
		\includegraphics[width=1\linewidth]{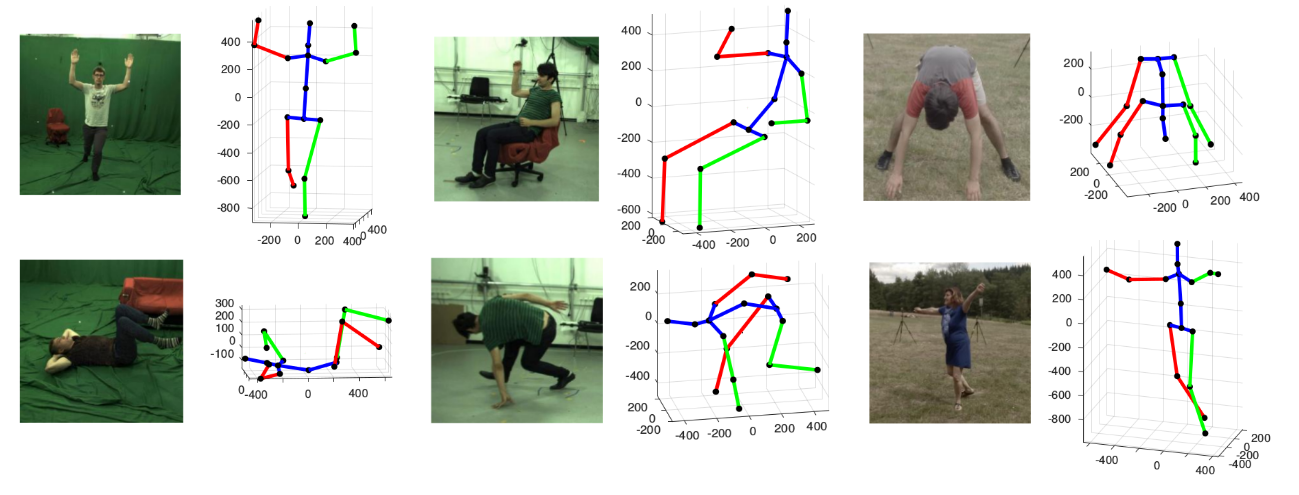}
	\end{center}
	\caption{Qualitative results on the MPI-INF-3DHP test set. Our model trained only on green-screen background images (see coloum 1) can still perform well on novel scenes (see column 2 and 3).}
	\label{fig:mpi-inf-3dhp}
\end{figure*}
    
\section{Conclusion}
In this paper, we propose a fully convolutional network that ties orientations with corresponding limb region to enhance the spatial relation between images and predictions. Our method is simple yet effective. Further experiments show that our model can generalize better to novel scenes and robust to inaccurate bounding boxes. In the future work, we expect adding additional constraints during both training and testing process would further improve the performance.

\newpage
\bibliography{egbib}

\begin{thebibliography}{38}
\providecommand{\natexlab}[1]{#1}
\providecommand{\url}[1]{\texttt{#1}}
\expandafter\ifx\csname urlstyle\endcsname\relax
  \providecommand{\doi}[1]{doi: #1}\else
  \providecommand{\doi}{doi: \begingroup \urlstyle{rm}\Url}\fi

\bibitem[Akhter and Black(2015)]{akhter2015pose}
Ijaz Akhter and Michael~J Black.
\newblock Pose-conditioned joint angle limits for 3d human pose reconstruction.
\newblock In \emph{CVPR}, 2015.

\bibitem[Andriluka et~al.(2014)Andriluka, Pishchulin, Gehler, and
  Schiele]{andriluka14cvpr}
Mykhaylo Andriluka, Leonid Pishchulin, Peter Gehler, and Bernt Schiele.
\newblock 2d human pose estimation: New benchmark and state of the art
  analysis.
\newblock In \emph{CVPR}, June 2014.

\bibitem[Cao et~al.(2017)Cao, Simon, Wei, and Sheikh]{cao2017realtime}
Zhe Cao, Tomas Simon, Shih-En Wei, and Yaser Sheikh.
\newblock Realtime multi-person 2d pose estimation using part affinity fields.
\newblock \emph{CVPR}, 2017.

\bibitem[Chen and Ramanan(2017)]{chen20163d}
Ching-Hang Chen and Deva Ramanan.
\newblock 3d human pose estimation= 2d pose estimation+ matching.
\newblock \emph{CVPR}, 2017.

\bibitem[Chu et~al.(2016{\natexlab{a}})Chu, Ouyang, Li, and
  Wang]{chu2016structured}
Xiao Chu, Wanli Ouyang, Hongsheng Li, and Xiaogang Wang.
\newblock Structured feature learning for pose estimation.
\newblock In \emph{CVPR}, 2016{\natexlab{a}}.

\bibitem[Chu et~al.(2016{\natexlab{b}})Chu, Ouyang, Wang, et~al.]{chu2016crf}
Xiao Chu, Wanli Ouyang, Xiaogang Wang, et~al.
\newblock Crf-cnn: Modeling structured information in human pose estimation.
\newblock In \emph{NIPS}, 2016{\natexlab{b}}.

\bibitem[Chu et~al.(2017)Chu, Yang, Ouyang, Ma, Yuille, and Wang]{chu2017multi}
Xiao Chu, Wei Yang, Wanli Ouyang, Cheng Ma, Alan Yuille, and Xiaogang Wang.
\newblock Multi-context attention for human pose estimation.
\newblock In \emph{CVPR}, 2017.

\bibitem[Collobert et~al.(2011)Collobert, Kavukcuoglu, and
  Farabet]{collobert2011torch7}
Ronan Collobert, Koray Kavukcuoglu, and Cl{\'e}ment Farabet.
\newblock Torch7: A matlab-like environment for machine learning.
\newblock In \emph{BigLearn, NIPS Workshop}, 2011.

\bibitem[Dabral et~al.(2018)Dabral, Mundhada, Kusupati, Afaque, Sharma, and
  Jain]{Dabral_2018_ECCV}
Rishabh Dabral, Anurag Mundhada, Uday Kusupati, Safeer Afaque, Abhishek Sharma,
  and Arjun Jain.
\newblock Learning 3d human pose from structure and motion.
\newblock In \emph{ECCV}, 2018.

\bibitem[Fang et~al.(2018)Fang, Xu, Wang, Liu, and Zhu]{fang2018learning}
Hao-Shu Fang, Yuanlu Xu, Wenguan Wang, Xiaobai Liu, and Song-Chun Zhu.
\newblock Learning pose grammar to encode human body configuration for 3d pose
  estimation.
\newblock In \emph{AAAI}, 2018.

\bibitem[Gong et~al.(2017)Gong, Liang, Shen, and Lin]{gong2017look}
Ke~Gong, Xiaodan Liang, Xiaohui Shen, and Liang Lin.
\newblock Look into person: Self-supervised structure-sensitive learning and a
  new benchmark for human parsing.
\newblock \emph{CVPR}, 2017.

\bibitem[Ionescu et~al.(2014)Ionescu, Papava, Olaru, and
  Sminchisescu]{h36m_pami}
Catalin Ionescu, Dragos Papava, Vlad Olaru, and Cristian Sminchisescu.
\newblock Human3.6m: Large scale datasets and predictive methods for 3d human
  sensing in natural environments.
\newblock \emph{IEEE Transactions on Pattern Analysis and Machine
  Intelligence}, 36\penalty0 (7):\penalty0 1325--1339, jul 2014.

\bibitem[Li and Chan(2014)]{li20143d}
Sijin Li and Antoni~B Chan.
\newblock 3d human pose estimation from monocular images with deep
  convolutional neural network.
\newblock In \emph{ACCV}. Springer, 2014.

\bibitem[Lin et~al.(2014)Lin, Maire, Belongie, Hays, Perona, Ramanan,
  Doll{\'a}r, and Zitnick]{lin2014microsoft}
Tsung-Yi Lin, Michael Maire, Serge Belongie, James Hays, Pietro Perona, Deva
  Ramanan, Piotr Doll{\'a}r, and C~Lawrence Zitnick.
\newblock Microsoft coco: Common objects in context.
\newblock In \emph{ECCV}, pages 740--755. Springer, 2014.

\bibitem[Martinez et~al.(2017)Martinez, Hossain, Romero, and
  Little]{martinez2017simple}
Julieta Martinez, Rayat Hossain, Javier Romero, and James~J Little.
\newblock A simple yet effective baseline for 3d human pose estimation.
\newblock \emph{ICCV}, 2017.

\bibitem[Mehta et~al.({\natexlab{a}})Mehta, Rhodin, Casas, Fua, Sotnychenko,
  Xu, and Theobalt]{mono-3dhp2017}
Dushyant Mehta, Helge Rhodin, Dan Casas, Pascal Fua, Oleksandr Sotnychenko,
  Weipeng Xu, and Christian Theobalt.
\newblock Monocular 3d human pose estimation in the wild using improved cnn
  supervision.
\newblock In \emph{3D Vision (3DV), 2017 Fifth International Conference on},
  {\natexlab{a}}.

\bibitem[Mehta et~al.({\natexlab{b}})Mehta, Sridhar, Sotnychenko, Rhodin,
  Shafiei, Seidel, Xu, Casas, and Theobalt]{VNect_SIGGRAPH2017}
Dushyant Mehta, Srinath Sridhar, Oleksandr Sotnychenko, Helge Rhodin, Mohammad
  Shafiei, Hans-Peter Seidel, Weipeng Xu, Dan Casas, and Christian Theobalt.
\newblock Vnect: Real-time 3d human pose estimation with a single rgb camera.
\newblock {\natexlab{b}}.

\bibitem[Moreno-Noguer(2017)]{moreno20163d}
Francesc Moreno-Noguer.
\newblock 3d human pose estimation from a single image via distance matrix
  regression.
\newblock \emph{CVPR}, 2017.

\bibitem[Newell et~al.(2016)Newell, Yang, and Deng]{newell2016stacked}
Alejandro Newell, Kaiyu Yang, and Jia Deng.
\newblock Stacked hourglass networks for human pose estimation.
\newblock In \emph{ECCV}, 2016.

\bibitem[Pavlakos et~al.(2017)Pavlakos, Zhou, Derpanis, and
  Daniilidis]{pavlakos2017coarse}
Georgios Pavlakos, Xiaowei Zhou, Konstantinos~G Derpanis, and Kostas
  Daniilidis.
\newblock Coarse-to-fine volumetric prediction for single-image 3d human pose.
\newblock \emph{CVPR}, 2017.

\bibitem[Pavlakos et~al.(2018)Pavlakos, Zhou, and
  Daniilidis]{pavlakos2018ordinal}
Georgios Pavlakos, Xiaowei Zhou, and Kostas Daniilidis.
\newblock Ordinal depth supervision for 3{D} human pose estimation.
\newblock In \emph{Computer Vision and Pattern Recognition (CVPR)}, 2018.

\bibitem[Ramakrishna et~al.(2012)Ramakrishna, Kanade, and
  Sheikh]{ramakrishna2012reconstructing}
Varun Ramakrishna, Takeo Kanade, and Yaser Sheikh.
\newblock Reconstructing 3d human pose from 2d image landmarks.
\newblock In \emph{ECCV}, 2012.

\bibitem[Rogez and Schmid(2016)]{rogez2016mocap}
Gr{\'e}gory Rogez and Cordelia Schmid.
\newblock Mocap-guided data augmentation for 3d pose estimation in the wild.
\newblock In \emph{NIPS}, 2016.

\bibitem[Ronchi et~al.(2018)Ronchi, Mac~Aodha, Eng, and
  Perona]{relativeposeBMVC18}
Matteo~Ruggero Ronchi, Oisin Mac~Aodha, Robert Eng, and Pietro Perona.
\newblock It's all relative: Monocular 3d human pose estimation from weakly
  supervised data.
\newblock In \emph{BMVC}, 2018.

\bibitem[Russakovsky et~al.(2015)Russakovsky, Deng, Su, Krause, Satheesh, Ma,
  Huang, Karpathy, Khosla, Bernstein, Berg, and Fei-Fei]{Russakovsky2015}
Olga Russakovsky, Jia Deng, Hao Su, Jonathan Krause, Sanjeev Satheesh, Sean Ma,
  Zhiheng Huang, Andrej Karpathy, Aditya Khosla, Michael Bernstein,
  Alexander~C. Berg, and Li~Fei-Fei.
\newblock Imagenet large scale visual recognition challenge.
\newblock \emph{International Journal of Computer Vision}, 115\penalty0
  (3):\penalty0 211--252, Dec 2015.

\bibitem[Sun et~al.(2017)Sun, Shang, Liang, and Wei]{sun2017compositional}
Xiao Sun, Jiaxiang Shang, Shuang Liang, and Yichen Wei.
\newblock Compositional human pose regression.
\newblock \emph{ICCV}, 2017.

\bibitem[Tekin et~al.(2017)Tekin, M{\'a}rquez-Neila, Salzmann, and
  Fua]{tekin2016fusing}
Bugra Tekin, Pablo M{\'a}rquez-Neila, Mathieu Salzmann, and Pascal Fua.
\newblock Fusing 2d uncertainty and 3d cues for monocular body pose estimation.
\newblock \emph{ICCV}, 2017.

\bibitem[Tome et~al.(2017)Tome, Russell, and Agapito]{tome2017lifting}
Denis Tome, Chris Russell, and Lourdes Agapito.
\newblock Lifting from the deep: Convolutional 3d pose estimation from a single
  image.
\newblock \emph{CVPR}, 2017.

\bibitem[Toshev and Szegedy(2014)]{toshev2014deeppose}
Alexander Toshev and Christian Szegedy.
\newblock Deeppose: Human pose estimation via deep neural networks.
\newblock In \emph{CVPR}, 2014.

\bibitem[Wang et~al.(2014)Wang, Wang, Lin, Yuille, and Gao]{wang2014robust}
Chunyu Wang, Yizhou Wang, Zhouchen Lin, Alan~L Yuille, and Wen Gao.
\newblock Robust estimation of 3d human poses from a single image.
\newblock In \emph{CVPR}, 2014.

\bibitem[Wei et~al.(2016)Wei, Ramakrishna, Kanade, and
  Sheikh]{wei2016convolutional}
Shih-En Wei, Varun Ramakrishna, Takeo Kanade, and Yaser Sheikh.
\newblock Convolutional pose machines.
\newblock In \emph{CVPR}, 2016.

\bibitem[Xia et~al.(2016)Xia, Wang, Chen, and Yuille]{xia2016zoom}
Fangting Xia, Peng Wang, Liang-Chieh Chen, and Alan~L Yuille.
\newblock Zoom better to see clearer: Human and object parsing with
  hierarchical auto-zoom net.
\newblock In \emph{ECCV}, 2016.

\bibitem[Yasin et~al.(2016)Yasin, Iqbal, Kruger, Weber, and
  Gall]{yasin2016dual}
Hashim Yasin, Umar Iqbal, Bjorn Kruger, Andreas Weber, and Juergen Gall.
\newblock A dual-source approach for 3d pose estimation from a single image.
\newblock In \emph{CVPR}, 2016.

\bibitem[Zhou et~al.(2015)Zhou, Leonardos, Hu, and Daniilidis]{zhou20153d}
Xiaowei Zhou, Spyridon Leonardos, Xiaoyan Hu, and Kostas Daniilidis.
\newblock 3d shape estimation from 2d landmarks: A convex relaxation approach.
\newblock In \emph{CVPR}, 2015.

\bibitem[Zhou et~al.(2016{\natexlab{a}})Zhou, Zhu, Leonardos, Derpanis, and
  Daniilidis]{zhou2016sparseness}
Xiaowei Zhou, Menglong Zhu, Spyridon Leonardos, Konstantinos~G Derpanis, and
  Kostas Daniilidis.
\newblock Sparseness meets deepness: 3d human pose estimation from monocular
  video.
\newblock In \emph{CVPR}, 2016{\natexlab{a}}.

\bibitem[Zhou et~al.(2017{\natexlab{a}})Zhou, Zhu, Pavlakos, Leonardos,
  Derpanis, and Daniilidis]{zhou2017monocap}
Xiaowei Zhou, Menglong Zhu, Georgios Pavlakos, Spyridon Leonardos,
  Kostantinos~G Derpanis, and Kostas Daniilidis.
\newblock Monocap: Monocular human motion capture using a cnn coupled with a
  geometric prior.
\newblock \emph{arXiv preprint arXiv:1701.02354}, 2017{\natexlab{a}}.

\bibitem[Zhou et~al.(2016{\natexlab{b}})Zhou, Sun, Zhang, Liang, and
  Wei]{zhou2016deep}
Xingyi Zhou, Xiao Sun, Wei Zhang, Shuang Liang, and Yichen Wei.
\newblock Deep kinematic pose regression.
\newblock In \emph{ECCV 2016 Workshops}, 2016{\natexlab{b}}.

\bibitem[Zhou et~al.(2017{\natexlab{b}})Zhou, Huang, Sun, Xue, and
  Wei]{zhou2017weakly}
Xingyi Zhou, Qixing Huang, Xiao Sun, Xiangyang Xue, and Yichen Wei.
\newblock Weakly-supervised transfer for 3d human pose estimation in the wild.
\newblock \emph{ICCV}, 2017{\natexlab{b}}.

\end{thebibliography}
\end{document}